\documentclass[USenglish]{article}	

\usepackage[utf8]{inputenc}				
\usepackage[big,online]{dgruyter}	
\usepackage{lmodern} 
\usepackage{microtype}
\usepackage[numbers,square,sort&compress]{natbib}

\usepackage{graphicx}
\usepackage{subcaption}
\begin{document}

\title{A Comparative Study on\\ Multi-task Uncertainty Quantification in \\Semantic Segmentation and \\Monocular Depth Estimation}
\runningtitle{A Comparative Study on Multi-task Uncertainties}

\author*[1]{Steven Landgraf}
\author[2]{Markus Hillemann}
\author[2]{Theodor Kapler} 
\author[2]{Markus Ulrich}
\runningauthor{S.~Landgraf et al.}
\affil[1]{\protect\raggedright 
Institute of Photogrammetry and Remote Sensing, Karlsruhe Institute of Technology, 76131 Karlsruhe, Germany, e-mail: steven.landgraf@kit.edu.}
\affil[2]{\protect\raggedright 
Institute of Photogrammetry and Remote Sensing, Karlsruhe Institute of Technology, 76131 Karlsruhe, Germany, e-mail: markus.hillemann@kit.edu, theodor.kapler@student.kit.edu, markus.ulrich@kit.edu. \newline \textbf{Extended Version:} This manuscript is an extended version of a previously published conference paper \cite{landgraf2024evaluation}.}

	
\abstract{Deep neural networks excel in perception tasks such as semantic segmentation and monocular depth estimation, making them indispensable in safety-critical applications like autonomous driving and industrial inspection. However, they often suffer from overconfidence and poor explainability, especially for out-of-domain data. While uncertainty quantification has emerged as a promising solution to these challenges, multi-task settings have yet to be explored. In an effort to shed light on this, we evaluate Monte Carlo Dropout, Deep Sub-Ensembles, and Deep Ensembles for joint semantic segmentation and monocular depth estimation. Thereby, we reveal that Deep Ensembles stand out as the preferred choice, particularly in out-of-domain scenarios, and show the potential benefit of multi-task learning with regard to the uncertainty quality in comparison to solving both tasks separately. Additionally, we highlight the impact of employing different uncertainty thresholds to classify pixels as certain or uncertain, with the median uncertainty emerging as a robust default.}


\maketitle

\section{Introduction} 
Deep neural networks are increasingly being used in real-time and safety-critical applications like autonomous driving \cite{mcallister2017ConcreteProblems}, industrial inspection 
\cite{steger2018MachineVision},
and automation \cite{landgraf2023segmentation}. Although they achieve unparalleled performance in fundamental perception tasks like semantic segmentation \cite{minaee2020ImageSegmentation} or monocular depth estimation \cite{dong2022towards}, they still suffer from problems like overconfidence \cite{guo2017CalibrationModerna}, lack explainability \cite{gawlikowski2022SurveyUncertainty}, and struggle to distinguish between in-domain and out-of-domain samples \cite{lee2018TrainingConfidencecalibrated}. 

In order to tackle these critical challenges and prevailing shortcomings of deep neural networks, a number of promising uncertainty quantification methods \cite{mackay1992PracticalBayesian,gal2016DropoutBayesian,lakshminarayanan2017SimpleScalable,valdenegro2023sub} have been proposed. Surprisingly, however, quantifying predictive uncertainties in the context of joint semantic segmentation and monocular depth estimation has not been thoroughly explored yet \cite{landgraf2024efficient}. Since many real-world applications are multi-modal in nature and, hence, have the potential to benefit from multi-task learning, this is a substantial gap in current literature.

To this end, we conduct a comprehensive series of experiments to study how multi-task learning influences the quality of uncertainty estimates in comparison to solving both tasks separately. Our contributions can be summarized as follows:
\begin{itemize}
    \itemsep0em 
    \item We combine three different uncertainty quantification methods - Monte Carlo Dropout (MCD), Deep Sub-Ensembles (DSE), and Deep Ensembles (DE) - with joint semantic segmentation and monocular depth estimation and evaluate their performance.
    \item Moreover, we study the impact of the number of ensemble members on the predictive performance and uncertainty quality.  
    \item In addition, we reveal the potential benefit of multi-task learning with regard to the uncertainty quality compared to solving semantic segmentation and monocular depth estimation separately. 
    \item Furthermore, we examine the influence of employing different uncertainty thresholds to determine whether a pixel is certain or uncertain.
    \item Finally, we show that Deep Ensembles exhibit robustness in out-of-domain scenarios, offering superior predictive performance and uncertainty quality over the baseline.
\end{itemize}

\section{Related Work}
In this section, we briefly summarize the related work on joint semantic segmentation and monocular depth estimation as well as uncertainty quantification.

\subsection{Joint Semantic Segmentation and Monocular Depth Estimation}
Semantic segmentation and monocular depth estimation are both essential tasks in image understanding, requiring pixel-wise predictions from a single input image. Due to the strong correlation and complementary nature of these tasks, several previous works have focused on addressing them jointly \cite{wang2015towards,mousavian2016joint,jiao2018look,xu2018pad,liu2018collaborative,lin2019depth,nekrasov2019real,he2021sosd,gao2022ci,ji2023semantic,kendall2018multi,liu2019end,bruggemann2021exploring,xu2022mtformer,bruggemann2020automated}. Notably, almost all previous works employ out-of-date architectures and require complex adaptations to either the model, the training process, or both. Instead of following this trend, we adapt a modern Vision-Transformer-based architecture similar to Xu et al. \cite{xu2022mtformer}, achieving competitive predictive performance while maintaining simplicity and transparency of the results.

\subsection{Uncertainty Quantification}
In order to address the shortcomings of deep neural networks, a variety of uncertainty quantification methods \cite{mackay1992PracticalBayesian,gal2016DropoutBayesian,lakshminarayanan2017SimpleScalable,valdenegro2023sub} and studies \cite{landgraf2024uce,wursthorn2024uq,wolf2024decoupling} have been published. The predictive uncertainty can be decomposed into aleatoric and epistemic uncertainty \cite{gal2016uncertainty}, which can be essential for applications like active learning and detecting out-of-distribution samples \cite{gal2017deep}. The aleatoric component captures the irreducible data uncertainty, such as image noise or noisy labels from imprecise measurements. The epistemic uncertainty accounts for the model uncertainty and can be reduced with more or higher quality training data \cite{gal2016uncertainty,kendall2017CVUncertainties}. 

Remarkably, quantifying uncertainties in joint semantic segmentation and monocular depth estimation has been largely overlooked \cite{landgraf2024efficient}. Therefore, we compare multiple uncertainty quantification methods for this task and show how multi-task learning influences the quality of the uncertainty quality in comparison to solving both tasks separately. 

\section{Evaluation Strategy}
In the following, we provide an overview of the baseline models and uncertainty quantification methods that are used in this comparative study.

\subsection{Baseline Models} To explore the impact of multi-task learning on the uncertainty quality, we conduct our evaluations with three models:
\begin{enumerate}
    \itemsep0em     
    \item SegFormer \cite{xie2021segformer} for the segmentation task,
    \item DepthFormer for the depth estimation task,
    \item SegDepthFormer for joint semantic segmentation and monocular depth estimation.
\end{enumerate}

\textbf{SegFormer.} For solving the semantic segmentation task by itself, we use SegFormer \cite{xie2021segformer}, a modern Transformer-based architecture. Due to its high efficiency and performance, it is particularly suitable for real-time applications that might rely on uncertainty quantification. We train all SegFormer models with the categorical Cross-Entropy loss
\begin{equation}
    \mathcal{L}_\mathrm{CE} = - \frac{1}{N} \sum_{n=1}^{N} \sum_{c=1}^{C} y_{n,c} \cdot \log(p(z)_{n, c})\enspace,
\end{equation}
where $N$ is the number of pixels in the image, $C$ is the number of classes, $y_{n,c}$ is the corresponding ground truth label, and $p(z)_{n,c}$ is the predicted softmax probability. The above loss only describes the calculation for a single image.

To obtain a measure for the aleatoric uncertainty \cite{kendall2017CVUncertainties} of the baseline model, we compute the predictive Entropy
\begin{equation}\label{eq: entropy}
    H(p(z)) = -\sum_{c=1}^{C} p(z)_c \cdot \log(p(z)_c)
    \enspace.
\end{equation}

\textbf{DepthFormer.} Highly inspired by the efficiency and performance of SegFormer \cite{xie2021segformer}, we propose DepthFormer for monocular depth estimation. We use the same hierarchical Transformer-based encoder and all-MLP decoder. In contrast to SegFormer, the output layer differs by having two output channnels: one for the predictive mean $\mu(z)$ and one for the predictive variance $s^2(z)$ \cite{loquercio2020general}. The first output channel uses a ReLU output activation function, while the second output channel applies Softplus activation, which is a smooth approximation of the ReLU functon with the advantage of being differentiable at $z = 0$. We found Softplus to work better than ReLU for the predictive variance, following the work of Lakshminarayanan et al. \cite{lakshminarayanan2017SimpleScalable}. 

For all DepthFormer models, we follow Nix and Weigend \cite{nix1994estimating} and treat the output of the model as a sample from a Gaussian distribution with the predictive mean $\mu(z)$ and a corresponding predictive variance $s^2(z)$. Based on this, we can minimize the Gaussian Negative Log-Likelihood (GNLL) loss for a single image with: 
\begin{equation}\label{eq: gnll}
    \mathcal{L}_\mathrm{GNLL} = \frac{1}{2} \left( \frac{(y - \mu(z))^2}{s^2(z)} + \log(s^2(z)) \right)
    \enspace,
\end{equation}
where $y$ is the ground truth depth. 

Through GNLL minimization, DepthFormer inherently learns corresponding variances, which can be interpreted as the aleatoric uncertainty \cite{kendall2017CVUncertainties,loquercio2020general}.

\begin{figure*}[htpb]
    \centering
    \includegraphics[width=0.99\linewidth]{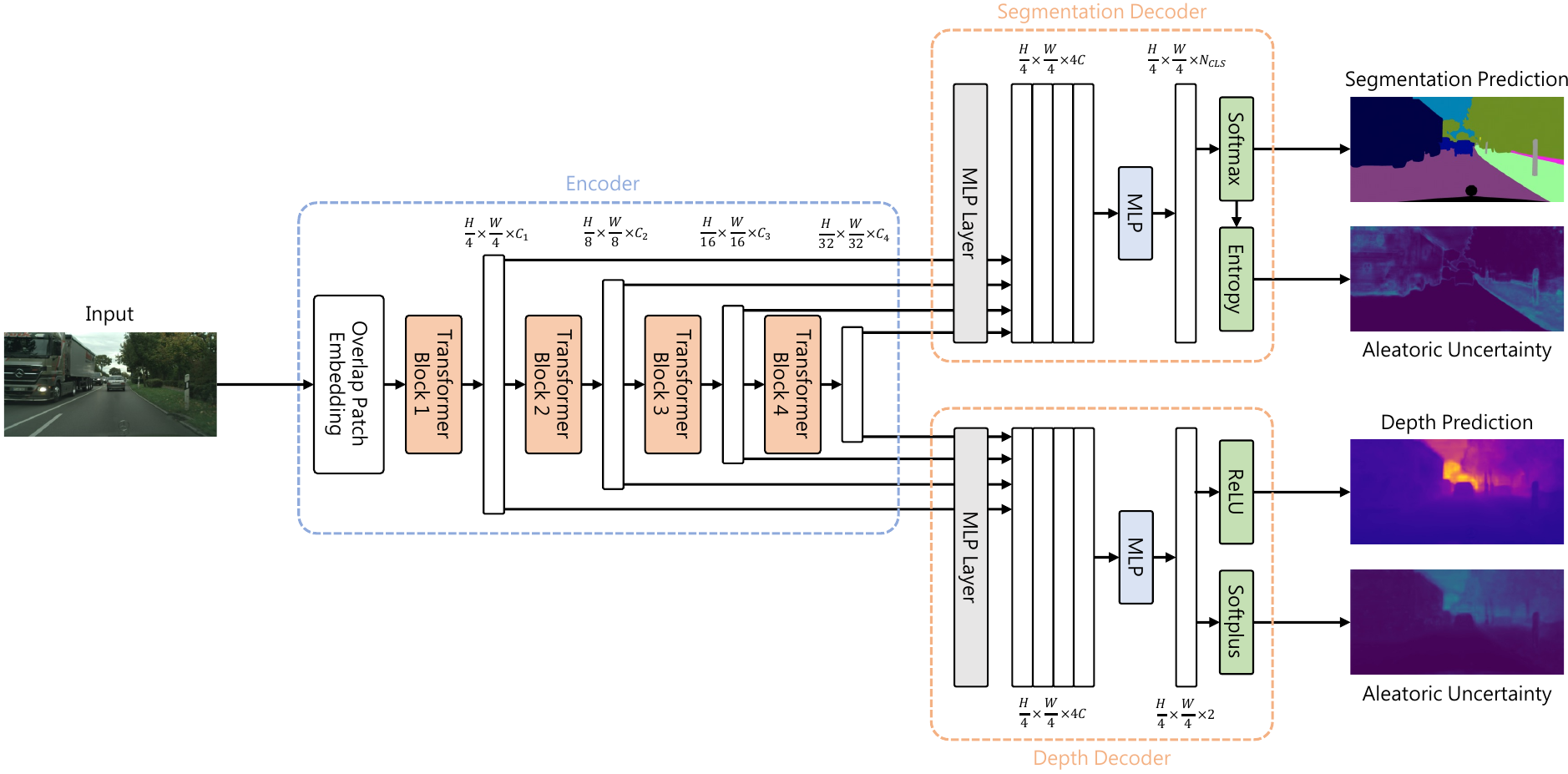}
    \caption{A schematic overview of the SegDepthFormer architecture. It combines the SegFormer \cite{xie2021segformer} architecture with a lightweight all-MLP depth decoder.}
    \label{fig:SegDepthFormer}
\end{figure*}

\textbf{SegDepthFormer.} To jointly solve semantic segmentation and monocular depth estimation, we propose SegDepthFormer. The architecture, which is shown in Figure \ref{fig:SegDepthFormer}, combines SegFormer \cite{xie2021segformer} and DepthFormer. It comprises three modules: a hierarchical Transformber-based encoder, an all-MLP segmentation decoder, and an all-MLP depth decoder. Both decoders fuse the multi-level features obtained through the shared encoder to solve the joint prediction task. 

SegDepthFormer is trained to minimize the weighted sum of the two previously described objective functions:
$ \mathcal{L} = \mathcal{L}_\mathrm{CE} + w_1 \mathcal{L}_\mathrm{GNLL}$, where $w_1$ is a weighting factor, which we set to $w_1 = 1$ for the sake of simplicity and because both loss values are of similar magnitude.

The respective aleatoric uncertainty is obtained by computing the predictive entropy $H(p(z))$ for the segmentation task or by the predictive variance $s^2(z)$, which is learned implicitly through the optimization of $\mathcal{L}_\mathrm{GNLL}$.

\subsection{Uncertainty Quantification}
We evaluate Monte Carlo Dropout (MCD) \cite{gal2016DropoutBayesian}, Deep Ensembles (DEs) \cite{lakshminarayanan2017SimpleScalable}, and Deep Sub-Ensembles (DSEs) \cite{valdenegro2023sub}, motivated by their simplicity, ease of implementation, parallelizability, minimal tuning requirements, and state-of-the-art performance. 

\textbf{Monte Carlo Dropout.} MCD depends on the number and placement of dropout layers, and particularly the dropout rate. We adopt the original SegFormer \cite{xie2021segformer} layer placement and consider two dropout rates, 20\% and 50\%. We sample ten times to obtain the prediction and predictive uncertainty \cite{gal2016DropoutBayesian, gustafsson2020evaluating}.

\textbf{Deep Ensemble.} DEs achieve the best results if they are trained to explore diverse modes in function space, which we accomplish by randomly initializing all decoder heads, using random augmentations, and by applying random shuffling of the training data points \cite{lakshminarayanan2017SimpleScalable, fort2020DeepEnsembles}. We report results of a DE with ten members, following the suggestions of previous work \cite{lakshminarayanan2017SimpleScalable, fort2020DeepEnsembles, landgraf2023dudes}.

\textbf{Deep Sub-Ensemble.} Consistent with DEs and MCD, we train the DSE with ten decoder heads for each task on top of a shared encoder \cite{valdenegro2023sub}. During training, we only optimize a single decoder head per training batch and alternate between them. Thereby, we aim to introduce as much randomness as possible, analogous to the training of DEs. For inference, we utilize all decoder heads. 

\section{Experimental Setup}
\textbf{Predictions.} For the semantic segmentation task, we compute the mean softmax probability of all samples. For the monocular depth estimation task, we first apply ReLU and then compute the mean depth of the corresponding samples.

\textbf{Uncertainty.} For the segmentation task, we compute the predictive entropy based on the mean softmax probabilities as a measure for the predictive uncertainty \cite{mukhoti2018evaluating}. For the depth estimation task, however, we calculate the predictive uncertainty based on the mean predictive variance and the variance of the depth predictions of the samples \cite{loquercio2020general}. 

\textbf{Datasets.} We conduct all experiments on Cityscapes \cite{cordts2016CityscapesDataset} and NYUv2 \cite{silberman2012indoor}. Cityscapes, with 2975 training and 500 validation images, is a popular urban street scene benchmark dataset. Notably, the depth values are based on the disparity of stereo camera images. NYUv2 contains 795 training and 654 testing images of indoor scenes.

\textbf{Data Augmentations.} Regardless of the trained model, we apply random scaling with a factor between $0.5$ and $2.0$, random cropping with a crop size of $768\times768$ pixels on Cityscapes and $480 \times 640$ pixels on NYUv2, and random horizontal flipping with a flip chance of $50\%$.

\textbf{Implementation Details.} For all training processes, we use AdamW \cite{loshchilov2017decoupled} optimizer with a base learning rate of $6\cdot10^{-5}$ and employ a polynomial rate scheduler:
\begin{equation}
lr = lr_\mathrm{base} \cdot (1 - \frac{\mathrm{iteration}}{\mathrm{total\:iterations}})^{0.9}
\enspace,
\end{equation}
where $lr$ is the current learning rate and $lr_{base}$ is the initial base learning rate. Besides, we use a batch size of 8 and train for 250 epochs on Cityscapes and for 100 epochs NYUv2, respectively. 
The encoders of the baseline models are initialized with weights pre-trained on ImageNet \cite{deng2009ImageNetLargescale} and then trained for 250 epochs on Cityscapes and for 100 epochs on NYUv2, respectively. We use the SegFormer-B2 \cite{xie2021segformer} backbone for all experiments.

\textbf{Metrics.} For semantic segmentation, we report mean Intersection over Union (mIoU) and Expected Calibration Error (ECE) \cite{naeini2015obtaining}. For monocular depth estimation, we use root mean squared error (RMSE). The uncertainty is evaluated using the following metrics proposed by Mukhoti and Gal \cite{mukhoti2018evaluating}:

\begin{enumerate}
    \itemsep0em     
    \item $p(\text{accurate}|\text{certain})$: The probability of accurate predictions given low uncertainty.
    \item $p(\text{uncertain}|\text{inaccurate})$: The probability of high uncertainty given inaccurate predictions.
    \item $PAvPU$: The combination of both cases, i.e., accurate$|$certain and inaccurate$|$uncertain. 
\end{enumerate}

Although these metrics have originally been proposed for semantic segmentation \cite{mukhoti2018evaluating}, we also use them to evaluate the depth uncertainty. We use the following formula to determine whether a depth prediction is accurate: 
\begin{equation}
    \max\left(\frac{\mu(z)}{y}, \frac{y}{\mu(z)}\right) = \delta_{1} < 1.25
    \enspace,
\end{equation}
where $\mu(z)$ is the predicted depth value of a pixel and $y$ is the corresponding ground truth depth. 
$\delta_{1}$ serves as a standard metric for quantifying the accuracy of monocular depth estimation models, using $1.25$ as the threshold to determine whether a depth prediction is accurate or not. In contrast, $\delta_{2}$ and $\delta_{3}$ are less strict, typically utilizing thresholds of $1.25^2$ and $1.25^3$, respectively.

For the sake of simplicity and to simulate real-world employment, we set the uncertainty threshold to the mean uncertainty \cite{mukhoti2018evaluating} of a given image for all evaluations, unless noted otherwise. We also conduct a comparative analysis of different thresholds, including the median and a statistically robust approach, which does not use the normal standard deviation, but one that is resilient to outliers. In order to achieve this, as described by Steger et al. \cite{steger2018MachineVision}, we take the median uncertainty
\begin{equation}
    X_M = \mathrm{median}(X) \enspace.
\end{equation}

Subsequently, a robust measure of the standard deviation can be derived by:
\begin{equation}
    \sigma = \frac{\mathrm{median}{(|X - X_M|)}}{0.6745} \enspace, 
\end{equation}
where the correction factor in the denominator is chosen in such a way that, for normally distributed uncertainties, the standard deviation aligns with one of a normal distribution.

Finally, a robust threshold can be determined by using the median as a central value and extending it by a range proportional to the robust standard deviation, scaled by a suitable factor $f$ (e.g., $\pm 2$ to capture approximately 95\% of the values in a normal distribution): 
\begin{equation}\label{equation: robust threshold}
    \tau = X_M \pm f \cdot \sigma \enspace.
\end{equation}
It is worth mentioning, however, that using a negative scaling factor $f$ resulted in unstable evaluation results, as it led to the threshold being set to $\tau = 0$ in some cases.

\section{Experiments}
The following section describes a variety of experiments, including quantitative results, the impact of the number of ensemble members, the impact of the uncertainty threshold, and an out-of-domain evaluation.

\subsection{Quantitative Results}
In this section, we describe the results of our joint uncertainty evaluation quantitatively. We compare combinations of the baseline models SegFormer, DepthFormer, and SegDepthFormer with the uncertainty quantification methods MCD, DSE, and DEs.
Tables \ref{table: uq cityscapes} and \ref{table: uq nyuv2} contain a detailed comparison, primarily focusing on the uncertainty quality.

\begin{table}[ht]
\centering
\caption{Quantitative comparison on the Cityscapes dataset \cite{cordts2016CityscapesDataset} between the three baseline models paired with MCD, DSE, and DEs, respectively. Best results are marked in \textbf{bold}.}
\resizebox{1.0\textwidth}{!}{
\setlength\extrarowheight{1mm}
\begin{tabular}{l|l|ccccc|cccc|c}
& & \multicolumn{5}{c|}{Semantic Segmentation} & \multicolumn{4}{c|}{Monocular Depth Estimation} &  \\ \cline{3-11}
& & mIoU $\uparrow$ & ECE $\downarrow$ & p(acc$|$cer) $\uparrow$ & p(unc$|$inacc) $\uparrow$ & PAvPU $\uparrow$ & RMSE $\downarrow$ & p(acc$|$cer) $\uparrow$ & p(unc$|$inacc) $\uparrow$ & PAvPU $\uparrow$ & Inference Time [ms] \\ \hline \hline

Baseline & SegFormer & 0.772 & 0.033 & 0.882 & 0.395 & 0.797 & - & - & - & - & 17.90 $\pm$ 0.47 \\
& DepthFormer & - & - & - & - & - & 7.452 & 0.749 & 0.476 & 0.766 & 17.59 $\pm$ 0.82 \\
& SegDepthFormer & 0.738 & 0.028 & 0.913 & 0.592 & 0.826 & 7.536 & 0.745 & 0.472 & 0.762 & 22.04 $\pm$ 0.27 \\ \hline

MCD (20\%) & SegFormer & 0.759 & \textbf{0.007} & 0.883 & 0.424 & 0.780 & - & - & - & - & 177.13 $\pm$ 0.64 \\
& DepthFormer & - & - & - & - & - & 7.956 & 0.749 & 0.555 & 0.739 & 139.32 $\pm$ 0.78 \\
& SegDepthFormer & 0.738 & 0.020 & 0.911 & 0.592 & 0.803 & 7.370 & 0.761 & 0.523 & 0.757 & 202.23 $\pm$ 0.39 \\ \hline

MCD (50\%) & SegFormer & 0.662 & 0.028 & 0.883 & 0.485 & 0.760 & - & - & - & - & 176.98 $\pm$ 0.53 \\
& DepthFormer & - & - & - & - & - & 21.602 & 0.181 & 0.366 & 0.431 & 139.81 $\pm$ 1.20 \\
& SegDepthFormer & 0.640 & 0.021 & 0.906 & 0.616 & 0.782 & 8.316 & 0.733 & \textbf{0.558} & 0.723 & 203.82 $\pm$ 0.81 \\ \hline

DSE & SegFormer & 0.772 & 0.037 & 0.890 & 0.456 & 0.797 & - & - & - & - & 132.30 $\pm$ 3.16 \\
& DepthFormer & - & - & - & - & - & \textbf{7.036} & 0.762 & 0.467 & 0.772 & 91.82 $\pm$ 2.01 \\
& SegDepthFormer & 0.749 & 0.009 & \textbf{0.931} & \textbf{0.696} & \textbf{0.844} & 7.441 & 0.751 & 0.463 & 0.766 & 212.11 $\pm$ 8.44 \\ \hline

DE & SegFormer & \textbf{0.784} & 0.033 & 0.887 & 0.416 & 0.798 & - & - & - & - & 667.51 $\pm$ 2.89 \\
& DepthFormer & - & - & - & - & - & 7.222 & 0.759 & 0.486 & 0.771 & 626.79 $\pm$ 2.05 \\
& SegDepthFormer & 0.755 & 0.015 & 0.917 & 0.609 & 0.828 & 7.156 & \textbf{0.763} & 0.493 & \textbf{0.773} & 743.23 $\pm$ 32.95 \\ 
\end{tabular}
}
\label{table: uq cityscapes}
\end{table}

\begin{table}[ht]
\centering
\caption{Quantitative comparison on the NYUv2 dataset \cite{silberman2012indoor} between the three baseline models paired with MCD, DSE, and DEs, respectively. Best results are marked in \textbf{bold}.}
\resizebox{1.0\textwidth}{!}{
\setlength\extrarowheight{1mm}
\begin{tabular}{l|l|ccccc|cccc|c}
& & \multicolumn{5}{c|}{Semantic Segmentation} & \multicolumn{4}{c|}{Monocular Depth Estimation} &  \\ \cline{3-11}
& & mIoU $\uparrow$ & ECE $\downarrow$ & p(acc$|$cer) $\uparrow$ & p(unc$|$inacc) $\uparrow$ & PAvPU $\uparrow$ & RMSE $\downarrow$ & p(acc$|$cer) $\uparrow$ & p(unc$|$inacc) $\uparrow$ & PAvPU $\uparrow$ & Inference Time [ms] \\ \hline \hline

Baseline & SegFormer & 0.470 & 0.159 & 0.768 & 0.651 & \textbf{0.734} & - & - & - & - & 18.09 $\pm$ 0.41 \\ 
& DepthFormer & - & - & - & - & - & 0.554 & 0.786 & 0.449 & 0.610 & 17.51 $\pm$ 0.87 \\ 
& SegDepthFormer & 0.466 & 0.151 & 0.769 & 0.659 & 0.733 & 0.558 & 0.776 & 0.446 & 0.594 & 22.31 $\pm$ 0.23 \\ \hline

MCD (20\%) & SegFormer & 0.422 & 0.102 & 0.767 & 0.706 & 0.724 & - & - & - & - & 222.67 $\pm$ 0.61 \\ 
& DepthFormer  & - & - & - & - & - & 0.605 & 0.741 & 0.478 & 0.568 & 139.58 $\pm$ 052 \\ 
& SegDepthFormer  & 0.433 & 0.093 & 0.771 & 0.710 & 0.725 & 0.610 & 0.731 & 0.450 & 0.560 & 251.25 $\pm$ 0.81 \\ \hline

MCD (50\%) & SegFormer & 0.273 & 0.083 & 0.705 & \textbf{0.722} & 0.713 & - & - & - & - & 223.25 $\pm$ 0.82 \\ 
& DepthFormer  & - & - & - & - & - & 0.978 & 0.516 & \textbf{0.492} & 0.526 & 139.27 $\pm$ 0.69 \\ 
& SegDepthFormer  & 0.272 & 0.084 & 0.702 & 0.721 & 0.711 & 0.837 & 0.576 & 0.473 & 0.525 & 251.98 $\pm$ 0.60 \\ \hline

DSE & SegFormer & 0.469 & 0.092 & 0.776 & 0.681 & 0.726 & - & - & - & - & 180.42 $\pm$ 3.93 \\ 
& DepthFormer & - & - & - & - & - & 0.547 & 0.782 & 0.423 & 0.596 & 91.66 $\pm$ 0.26 \\ 
& SegDepthFormer & 0.461 & \textbf{0.077} & 0.776 & 0.692 & 0.723 & 0.584 & 0.738 & 0.403 & 0.573 & 261.69 $\pm$ 5.10 \\ \hline

DE & SegFormer & \textbf{0.486} & 0.125 & 0.782 & 0.675 & \textbf{0.734} & - & - & - & - & 715.97 $\pm$ 7.55 \\
& DepthFormer & - & - & - & - & - & \textbf{0.524} & \textbf{0.808} & 0.475 & \textbf{0.613} & 624.30 $\pm$ 2.07 \\ 
& SegDepthFormer & 0.481 & 0.122 & \textbf{0.783} & 0.682 & 0.733 & 0.552 & 0.785 & 0.453 & 0.590 & 788.76 $\pm$ 2.00 \\ 
\end{tabular}
}
\label{table: uq nyuv2}
\end{table}

\textbf{Single-task vs. Multi-task.} Looking at the differences between the single-task models, SegFormer and DepthFormer, and the multi-task model, SegDepthFormer, the single-task models generally deliver slightly better prediction performance. However, SegDepthFormer exhibits greater uncertainty quality for the semantic segmentation task in comparison to SegFormer. This is particularly evident for $p(\text{uncertain}|\text{inaccurate})$ on Cityscapes. For the depth estimation task, there is no significant difference in terms of uncertainty quality.

\textbf{Baseline Models.} As expected, the baseline models have the lowest inference times, being 5 to 30 times faster without using any uncertainty quantification method. While their prediction performance turns out to be quite competitive, only beaten by DEs, they show poor calibration and uncertainty quality for semantic segmentation. Surprisingly, the uncertainty quality for the depth estimation task is very decent, often only surpassed by the DE. 

\textbf{Monte Carlo Dropout.} MCD causes a significantly higher inference time compared to the respective baseline model. Additionally, leaving dropout activated during inference to sample from the posterior has a detrimental effect on the prediction performance, particularly with a 50\% dropout ratio. Nevertheless, MCD outputs well-calibrated softmax probabilities and uncertainties, although the results should be interpreted with caution because of the deteriorated prediction quality. 

\textbf{Deep Sub-Ensemble.} Across both datasets, DSEs show comparable prediction performance compared with the baseline models. Notably, DSEs consistently demonstrate a high uncertainty quality across all metrics, particularly in the segmentation task on Cityscapes.

\textbf{Deep Ensemble.} In accordance with previous work \cite{gustafsson2020evaluating}, DEs emerge as state-of-the-art, delivering the best prediction performance and mostly superior uncertainty quality. At the same time, DEs suffer from the highest computational cost, which scales approximately linearly with the number of members. Hence, we will explore the impact of ensemble members next.

\subsection{Impact of Ensemble Members}
This section explores the impact of the number of ensemble members in the SegDepthFormer Deep Ensemble on predictive performance and uncertainty quality.

\begin{figure}[htpb]
    \centering
    \begin{subfigure}[b]{0.47\textwidth}
        \includegraphics[width=\textwidth]{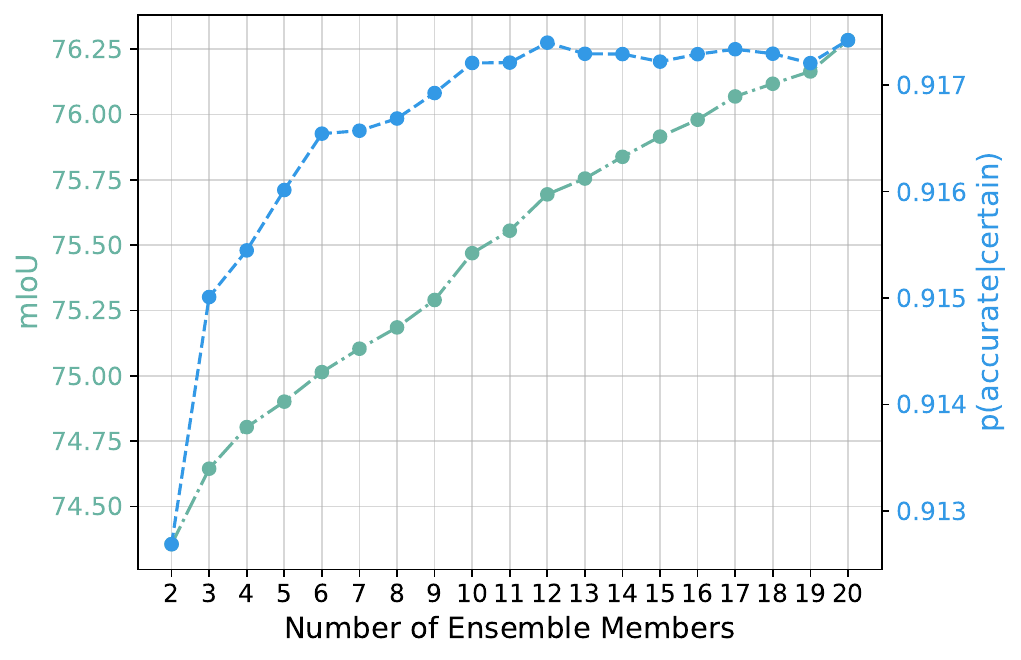}
        \caption{Segmentation: $p(\text{acc}|\text{cer})$}
        \label{fig:subfig1}
    \end{subfigure}
    \hfill
    \begin{subfigure}[b]{0.47\textwidth}
        \includegraphics[width=\textwidth]{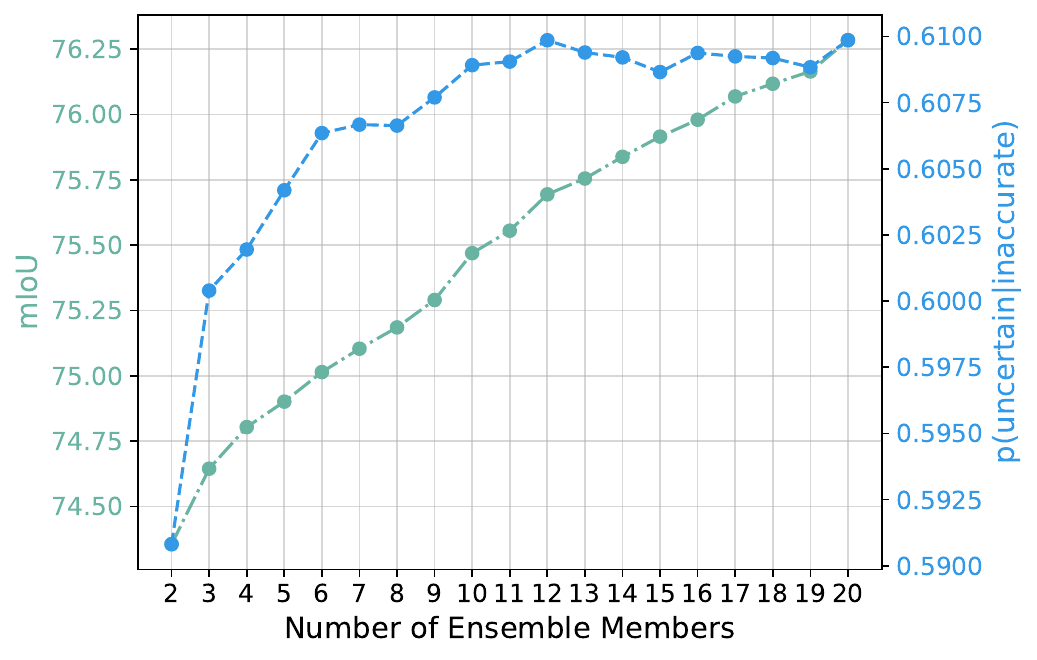}
        \caption{Segmentation: $p(\text{unc}|\text{inacc})$}
        \label{fig:subfig2}
    \end{subfigure}
    \hfill
    \begin{subfigure}[b]{0.47\textwidth}
        \includegraphics[width=\textwidth]{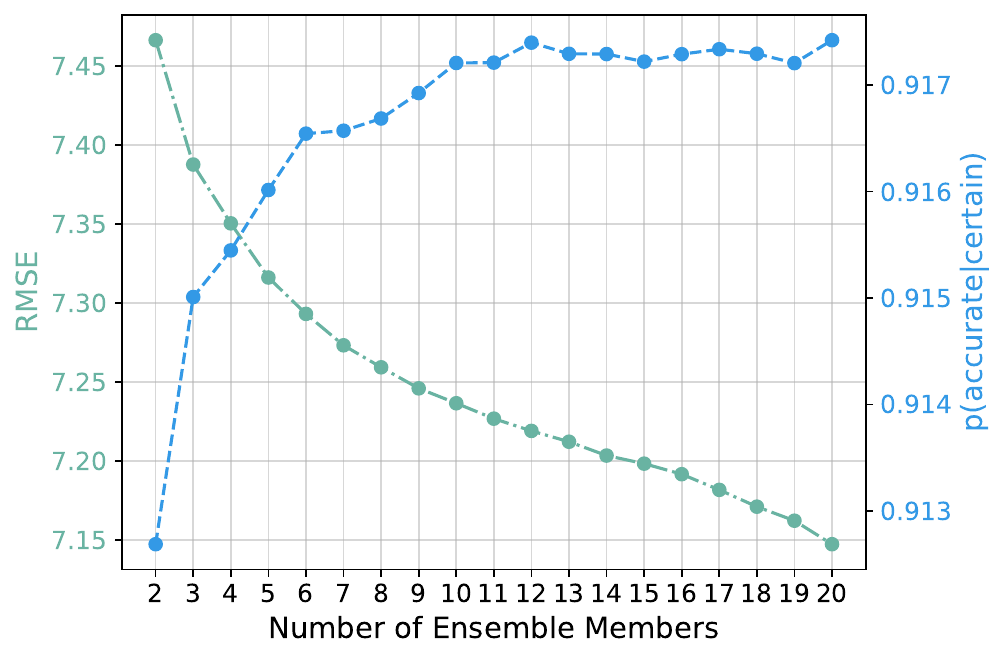}
        \caption{Depth: $p(\text{acc}|\text{cer})$}
        \label{fig:subfig3}
    \end{subfigure}
    \hfill
    \begin{subfigure}[b]{0.47\textwidth}
        \includegraphics[width=\textwidth]{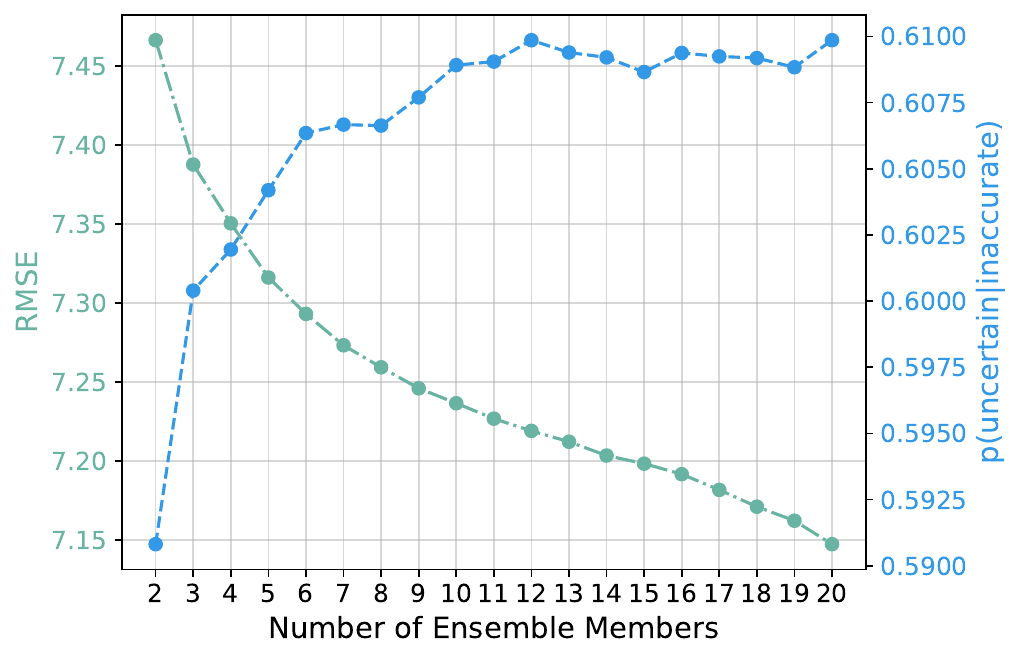}
        \caption{Depth: $p(\text{unc}|\text{inacc})$}
        \label{fig:subfig4}
    \end{subfigure}
    \caption{Impact of the number of ensemble members on the predictive performance and uncertainty quality for a SegDepthFormer Deep Ensemble on the Cityscapes dataset \cite{cordts2016CityscapesDataset}.}
    \label{fig:Ensemble_Member}
\end{figure}

As Figure \ref{fig:Ensemble_Member} shows, increasing the number of ensemble members improves the mIoU from 74.35\% with just two members to a maximum of just over 76.25\% with twenty members. Similarly, the RMSE decreases from 7.46 to just under 7.15. More notably, however, the uncertainty quality — measured by $p(\text{accurate}|\text{certain})$ and $p(\text{uncertain}|\text{inaccurate})$ — does not show significant improvement after 12 members for both tasks. Given the high computational cost of adding more members to a Deep Ensemble and the diminishing returns in both predictive performance but mainly uncertainty quality, a configuration of just over ten members appears to be a reasonable trade-off. These findings align with prior work on Deep Ensemble-based uncertainty quantification \cite{fort2020DeepEnsembles,landgraf2023dudes,landgraf2024uce}, where ten members are often recommended as the default choice.

\subsection{Impact of Uncertainty Threshold}\label{sec: UT Impact}
In this section, we examine the impact of employing different thresholds to classify pixels as certain or uncertain. In addition to the commonly used mean and median thresholds, we incorporate a robust alternative based on a standard deviation designed to mitigate the influence of outliers (see Eq. \ref{equation: robust threshold}) \cite{steger2018MachineVision}. Tables \ref{table: threshold cityscapes} and \ref{table: threshold nyuv2} provide a quantitative comparison of these three thresholding strategies, highlighting their effects on the uncertainty quality. 

\begin{table}[ht]
\centering
\caption{Uncertainty threshold comparison on the Cityscapes dataset \cite{cordts2016CityscapesDataset} between the SegDepthFormer baseline model and a SegDepthFormer DE. Best results are marked in \textbf{bold}.}
\resizebox{1.0\textwidth}{!}{
\setlength\extrarowheight{1mm}
\begin{tabular}{l|l|ccc|ccc}
& & \multicolumn{3}{c|}{Semantic Segmentation} & \multicolumn{3}{c}{Monocular Depth Estimation}  \\ \cline{3-8}
& & p(acc$|$cer) $\uparrow$ & p(unc$|$inacc) $\uparrow$ & PAvPU $\uparrow$ & p(acc$|$cer) $\uparrow$ & p(unc$|$inacc) $\uparrow$ & PAvPU $\uparrow$ \\ \hline \hline

Baseline & Mean   & 0.913 & 0.592 & 0.826 & 0.745 & 0.472 & 0.762 \\
         & Median & 0.947 & 0.852 & 0.611 & 0.870 & 0.832 & 0.750 \\
         & Robust \textsubscript{($f=1$)} & 0.939 & 0.806 & 0.667 & 0.810 & 0.685 & 0.769 \\
         & Robust \textsubscript{($f=2$)} & 0.935 & 0.780 & 0.693 & 0.779 & 0.596 & 0.766 \\ \hline
DE       & Mean   & 0.917 & 0.609 & \textbf{0.828} & 0.763 & 0.492 & \textbf{0.773} \\
         & Median & \textbf{0.950} & \textbf{0.861} & 0.612 & \textbf{0.878} & \textbf{0.834} & 0.743 \\
         & Robust \textsubscript{($f=1$)} & 0.943 & 0.816 & 0.670 & 0.822 & 0.691 & 0.770 \\
         & Robust \textsubscript{($f=2$)} & 0.939 & 0.790 & 0.698 & 0.791 & 0.596 & 0.770 \\
\end{tabular}
}
\label{table: threshold cityscapes}
\end{table}

\begin{table}[ht]
\centering
\caption{Uncertainty threshold comparison on the NYUv2 dataset \cite{silberman2012indoor} between the SegDepthFormer baseline model and a SegDepthFormer DE. Best results are marked in \textbf{bold}.}
\resizebox{1.0\textwidth}{!}{
\setlength\extrarowheight{1mm}
\begin{tabular}{l|l|ccc|ccc}
& & \multicolumn{3}{c|}{Semantic Segmentation} & \multicolumn{3}{c}{Monocular Depth Estimation}  \\ \cline{3-8}
& & p(acc$|$cer) $\uparrow$ & p(unc$|$inacc) $\uparrow$ & PAvPU $\uparrow$ & p(acc$|$cer) $\uparrow$ & p(unc$|$inacc) $\uparrow$ & PAvPU $\uparrow$ \\ \hline \hline

Baseline & Mean   & 0.769 & 0.659 & \textbf{0.733} & 0.776 & 0.446 & 0.594 \\
         & Median & 0.810 & 0.783 & 0.698 & \textbf{0.787} & \textbf{0.591} & 0.532 \\
         & Robust \textsubscript{($f=1$)} & 0.743 & 0.572 & 0.691 & 0.772 & 0.303 & 0.670 \\
         & Robust \textsubscript{($f=2$)} & 0.706 & 0.423 & 0.667 & 0.766 & 0.172 & 0.717 \\ \hline
DE       & Mean   & 0.783 & 0.682 & \textbf{0.733} & 0.785 & 0.453 & 0.590 \\
         & Median & \textbf{0.816} & \textbf{0.785} & 0.696 & 0.796 & 0.589 & 0.528 \\
         & Robust \textsubscript{($f=1$)} & 0.741 & 0.544 & 0.688 & 0.784 & 0.309 & 0.676 \\
         & Robust \textsubscript{($f=2$)} & 0.701 & 0.381 & 0.662 & 0.779 & 0.178 & \textbf{0.728} \\
\end{tabular}
}
\label{table: threshold nyuv2}
\end{table}

As shown by Tables \ref{table: threshold cityscapes} and \ref{table: threshold nyuv2}, the median threshold consistently performs best for $p(\text{accurate}|\text{certain})$ and $p(\text{uncertain}|\text{inaccurate})$, demonstrating its ability in correlating correct predictions with low uncertainty and incorrect predictions with high uncertainty across both tasks and datasets. However, for the combined uncertainty quality metric, PAvPU, the mean threshold often achieves the highest scores, indicating that it provides fewer accurate labels with high uncertainty than the median threshold. While the robust threshold is theoretically promising and provides higher scores for $p(\text{accurate}|\text{certain})$ and $p(\text{uncertain}|\text{inaccurate})$ than the mean threshold and higher scores for PAvPU than the median threshold on the Cityscapes dataset, its performance is less convincing on the NYUv2 dataset, particularly for $p(\text{uncertain}|\text{inaccurate})$. This suggests that the robust threshold may require further tuning of the clipping factor $f$ or more sophisticated adaptions to perform consistently across diverse datasets to match the reliability of the other two approaches. 

Overall, the choice of the uncertainty threshold significantly impacts the uncertainty quality metrics, influencing the correlation between accurate predictions with low uncertainty and inaccurate predictions with high uncertainty. However, the results indicate that this impact is consistent across different methods, suggesting that the threshold selection influences the metrics independently of the underlying model or approach used.

\subsection{Out-of-Domain Evaluation}
In the following, we analyze the predictive performance and uncertainty quality of the SegDepthFormer baseline model and a SegDepthFormer DE with 10 members on two out-of-domain (OOD) datasets: Foggy Cityscapes \cite{sakaridis2018semantic} and Rain Cityscapes \cite{hu2019depth}. Both models were originally trained on the standard Cityscapes dataset and are evaluated on the OOD datasets without fine-tuning. Quantitative comparisons for multiple variations of both datasets, involving different attenuation coefficients that control fog density or rain intensity, are presented in Tables \ref{table:uq-foggy} and \ref{table:uq-rain}.

\textbf{Foggy Cityscapes.} Compared to the original Cityscapes dataset, the Foggy Cityscapes dataset reveals significant performance degradation, as shown by Table \ref{table:uq-foggy}. For the strongest fog, the baseline model's performance drops more significantly from 0.738 to 0.609 mIoU and from 7.536 to 9.844 RMSE, while the DE model shows a decline from 0.755 to 0.639 mIoU and from 7.156 to 9.213 RMSE. Similarly, calibration quality decreases, with ECE increasing from 0.028 to 0.078 for the baseline and from 0.015 to 0.045 for the DE. Regarding the uncertainty quality, the baseline exhibits consistent deterioration in $p(\text{accurate}|\text{certain})$ and PAvPU with denser fog but maintains relatively stable $p(\text{uncertain}|\text{inaccurate})$, likely due to an increasing number of inaccurate pixels offsetting the apparent uncertainty degradation. The DE does not only demonstrate less pronounced performance degradation but also manages to improve $p(\text{uncertain}|\text{inaccurate})$ under severe fog conditions, reflecting its robustness in uncertainty quantification.

\begin{table}[ht]
\centering
\caption{Quantitative comparison of the SegDepthFormer baseline model and a SegDepthFormer DE with 10 members on the Foggy Cityscapes validation dataset \cite{sakaridis2018semantic} without fine-tuning. $\beta$ denotes the attenuation coefficient and controls the thickness of the fog. Higher $\beta$ values result in thicker fog. The original Cityscapes and the Foggy Cityscapes datasets share the same validation images, enabling a fair comparison between in-domain and out-of-domain results.}
\resizebox{1.0\textwidth}{!}{
\setlength\extrarowheight{1mm}
\begin{tabular}{l|l|ccccc|cccc}
& & \multicolumn{5}{c|}{Semantic Segmentation} & \multicolumn{4}{c}{Monocular Depth Estimation} \\ \cline{3-11}
& & mIoU $\uparrow$ & ECE $\downarrow$ & p(acc$|$cer) $\uparrow$ & p(unc$|$inacc) $\uparrow$ & PAvPU $\uparrow$ & RMSE $\downarrow$ & p(acc$|$cer) $\uparrow$ & p(unc$|$inacc) $\uparrow$ & PAvPU $\uparrow$ \\ \hline \hline

Cityscapes & Baseline & 0.738 & 0.028 & 0.913 & 0.592 & 0.826 & 7.536 & 0.745 & 0.472 & 0.762 \\ 
& DE & 0.755 & 0.015 & 0.917 & 0.609 & 0.828 & 7.156 & 0.763 & 0.493 & 0.773 \\ \hline \hline

Foggy$_{\beta=0.005}$ & Baseline & 0.707 & 0.035 & 0.906 & 0.602 & 0.818 & 8.061 & 0.731 & 0.481 & 0.751 \\ 
& DE & 0.727 & 0.028 & 0.914 & 0.627 & 0.822 & 7.487 & 0.758 & 0.509 & 0.765 \\ \hline

Foggy$_{\beta=0.01}$ & Baseline & 0.674 & 0.054 & 0.899 & 0.606 & 0.814 & 8.628 & 0.715 & 0.475 & 0.741 \\ 
& DE & 0.699 & 0.056 & 0.910 & 0.637 & 0.817 & 7.971 & 0.750 & 0.511 & 0.761 \\ \hline

Foggy$_{\beta=0.02}$ & Baseline & 0.609 & 0.078 & 0.875 & 0.593 & 0.798 & 9.844 & 0.697 & 0.467 & 0.730 \\ 
& DE & 0.639 & 0.045 & 0.895 & 0.644 & 0.803 & 9.213 & 0.738 & 0.517 & 0.760 \\ 
\end{tabular}
}
\label{table:uq-foggy}
\end{table}

\textbf{Rain Cityscapes.} Table \ref{table:uq-rain} highlights performance trends across varying levels of simulated rain. Under Rain$_1$, the baseline model achieves 0.608 mIoU and 7.187 RMSE, while the DE model improves these metrics to 0.673 mIoU and 6.740 RMSE. Calibration quality also favors the DE, with ECE values of 0.020 for the baseline and 0.004 for the DE. For uncertainty quality, $p(\text{accurate}|\text{certain})$, $p(\text{uncertain}|\text{inaccurate})$, and PAvPU consistently show better results for the DE. As rain intensity increases to Rain$_2$ and Rain$_3$, both models experience performance degradation, but the DE retains superior metrics, achieving 0.612 mIoU and 8.294 RMSE under Rain$_3$ compared to 0.582 mIoU and 8.848 RMSE for the baseline. Additionally, the DE sustains better calibration and uncertainty metrics, demonstrating enhanced robustness to OOD conditions.

\begin{table}[ht]
\centering
\caption{Quantitative comparison of the SegDepthFormer baseline model and a SegDepthFormer DE with 10 members on the Rain Cityscapes validation dataset \cite{hu2019depth} without fine-tuning. $\beta$ denotes the attenuation coefficient and controls the thickness of the fog. Higher $\beta$ values result in thicker fog. We evaluate on three sets of parameters, where Rain$_1$ uses [0.01, 0.005, 0.01], Rain$_2$ uses [0.02, 0.01, 0.005], and Rain$_3$ uses [0.03, 0.015, 0.002] for attenuation coefficients $\alpha$ and $\beta$ as well as the raindrop radius $a$. $\alpha$ and $\beta$ determine the degree of simulated rain and fog in the images.}
\resizebox{1.0\textwidth}{!}{
\setlength\extrarowheight{1mm}
\begin{tabular}{l|l|ccccc|cccc}
& & \multicolumn{5}{c|}{Semantic Segmentation} & \multicolumn{4}{c}{Monocular Depth Estimation} \\ \cline{3-11}
& & mIoU $\uparrow$ & ECE $\downarrow$ & p(acc$|$cer) $\uparrow$ & p(unc$|$inacc) $\uparrow$ & PAvPU $\uparrow$ & RMSE $\downarrow$ & p(acc$|$cer) $\uparrow$ & p(unc$|$inacc) $\uparrow$ & PAvPU $\uparrow$ \\ \hline \hline

Rain$_{1}$ & Baseline & 0.608 & 0.020 & 0.936 & 0.658 & 0.810 & 7.187 & 0.792 & 0.558 & 0.767 \\ 
& DE & 0.673 & 0.004 & 0.954 & 0.741 & 0.813 & 6.740 & 0.804 & 0.559 & 0.767 \\ \hline

Rain$_{2}$ & Baseline & 0.611 & 0.031 & 0.928 & 0.670 & 0.802 & 8.043 & 0.771 & 0.543 & 0.756 \\ 
& DE & 0.645 & 0.012 & 0.948 & 0.750 & 0.806 & 7.516 & 0.785 & 0.544 & 0.759 \\ \hline

Rain$_{3}$ & Baseline & 0.582 & 0.045 & 0.917 & 0.671 & 0.795 & 8.848 & 0.751 & 0.534 & 0.749 \\ 
& DE & 0.612 & 0.023 & 0.943 & 0.756 & 0.799 & 8.294 & 0.767 & 0.535 & 0.755 \\ 
\end{tabular}
}
\label{table:uq-rain}
\end{table}

\begin{figure}[htpb]
    \centering
    \begin{subfigure}[b]{0.31\textwidth}
        \includegraphics[width=\textwidth]{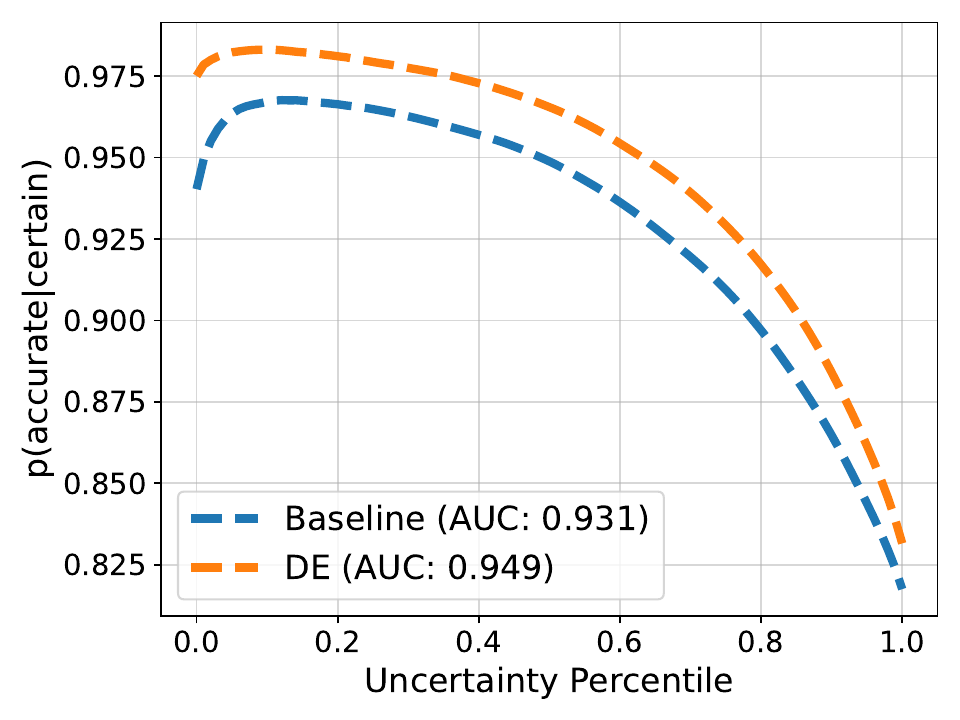}
        \caption{Seg.: $p(\text{acc}|\text{cer})$}
        \label{fig: subfig1}
    \end{subfigure}
    \hfill
    \begin{subfigure}[b]{0.31\textwidth}
        \includegraphics[width=\textwidth]{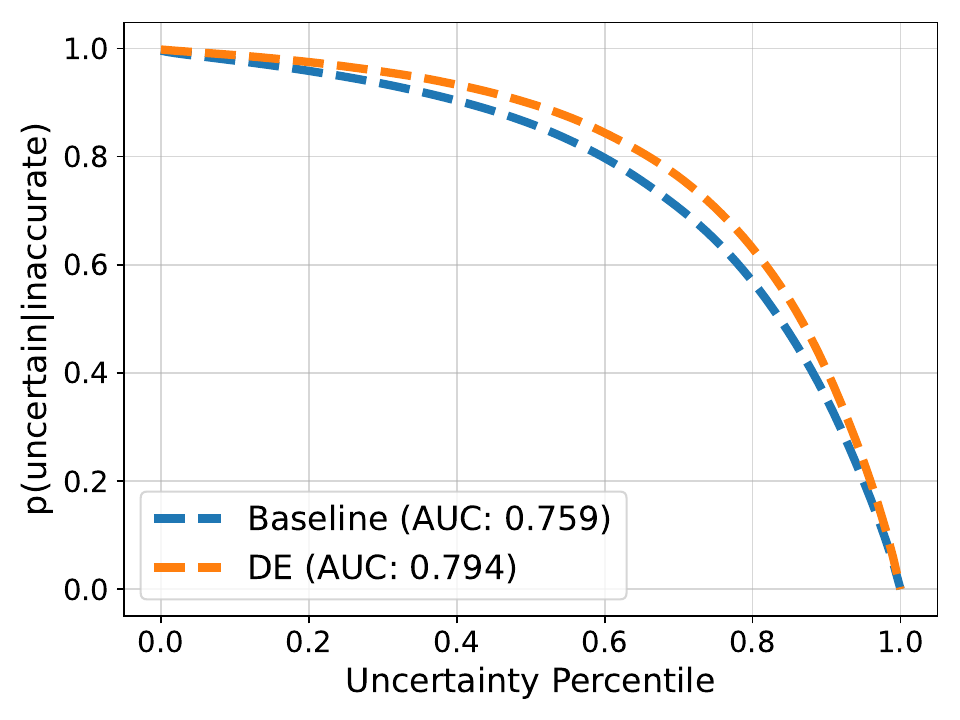}
        \caption{Seg.: $p(\text{unc}|\text{inacc})$}
        \label{fig: subfig2}
    \end{subfigure}
    \hfill
    \begin{subfigure}[b]{0.31\textwidth}
        \includegraphics[width=\textwidth]{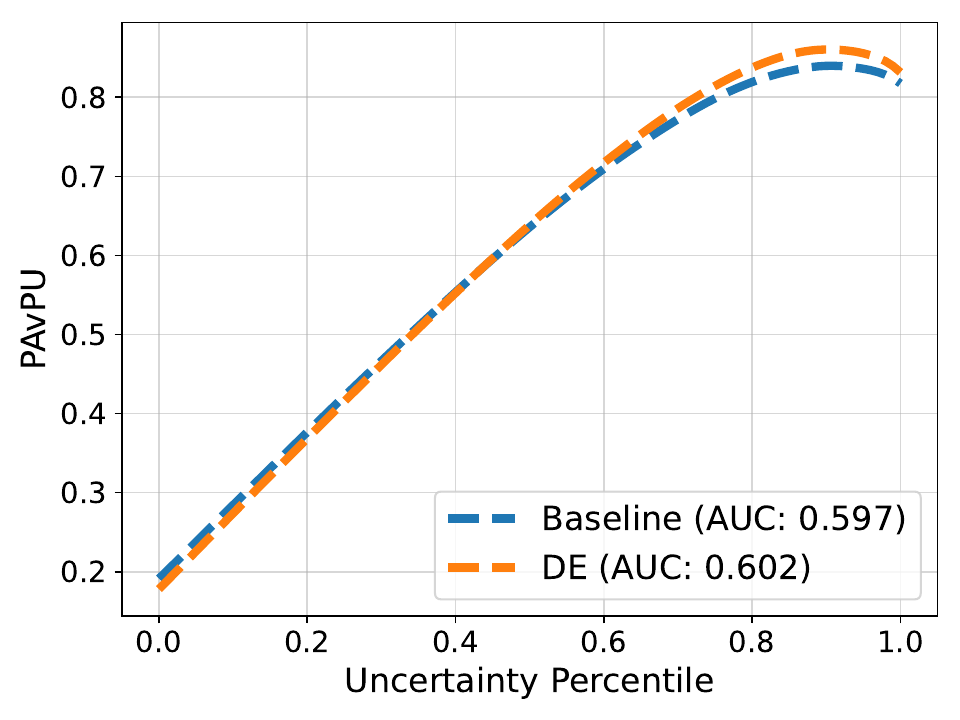}
        \caption{Seg.: PAvPU}
        \label{fig: subfig3}
    \end{subfigure}
    \hfill
    \begin{subfigure}[b]{0.31\textwidth}
        \includegraphics[width=\textwidth]{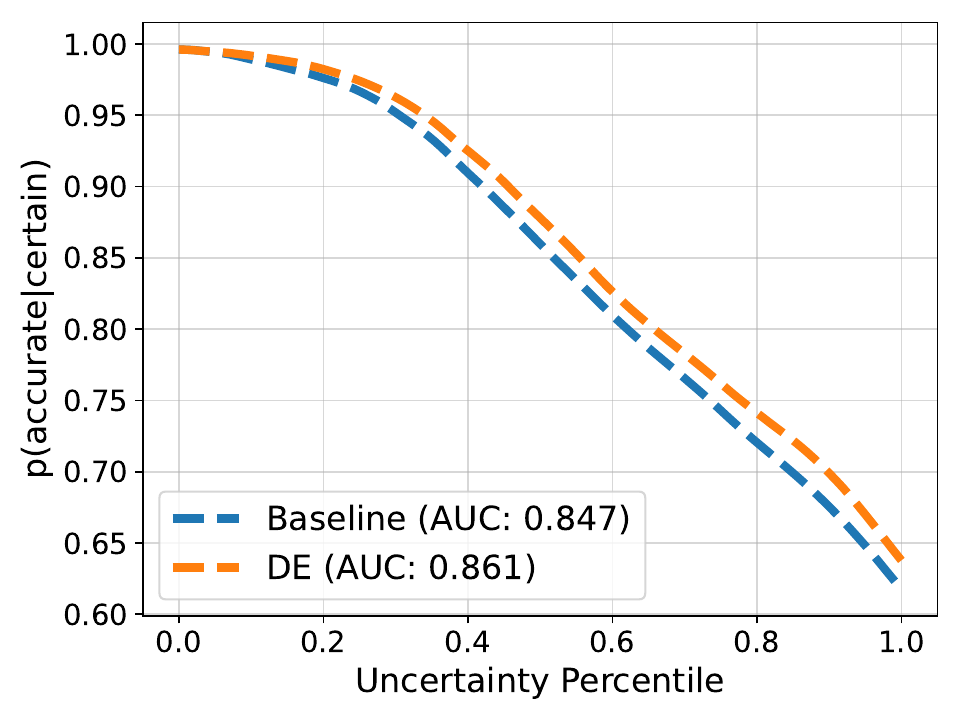}
        \caption{Depth: $p(\text{acc}|\text{cer})$}
        \label{fig: subfig4}
    \end{subfigure}
    \hfill
    \begin{subfigure}[b]{0.31\textwidth}
        \includegraphics[width=\textwidth]{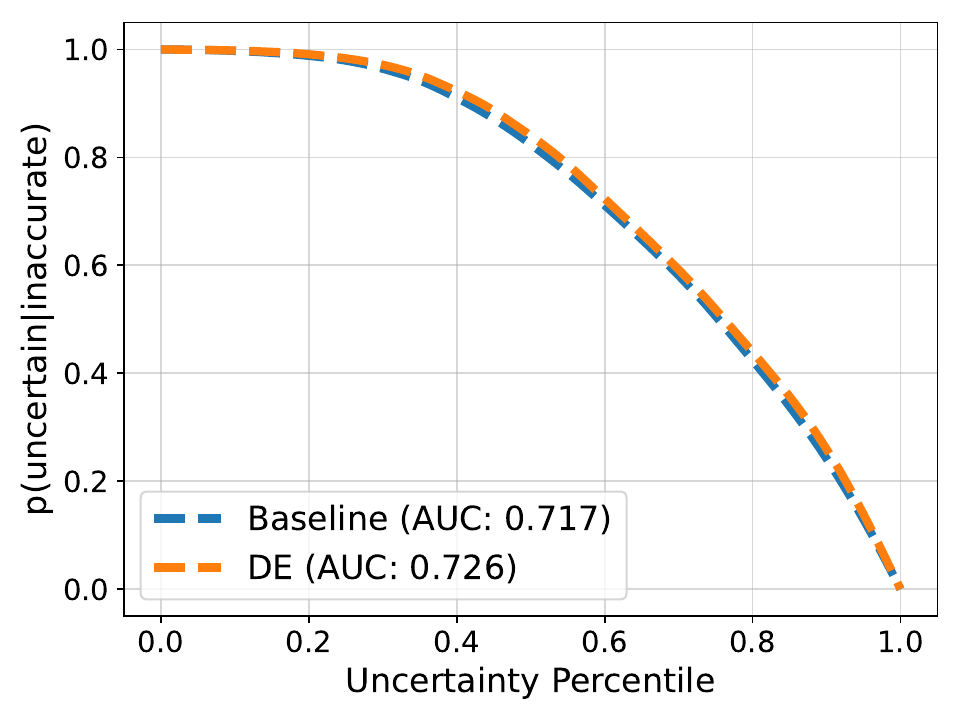}
        \caption{Depth: $p(\text{unc}|\text{inacc})$}
        \label{fig: subfig5}
    \end{subfigure}
    \hfill
    \begin{subfigure}[b]{0.31\textwidth}
        \includegraphics[width=\textwidth]{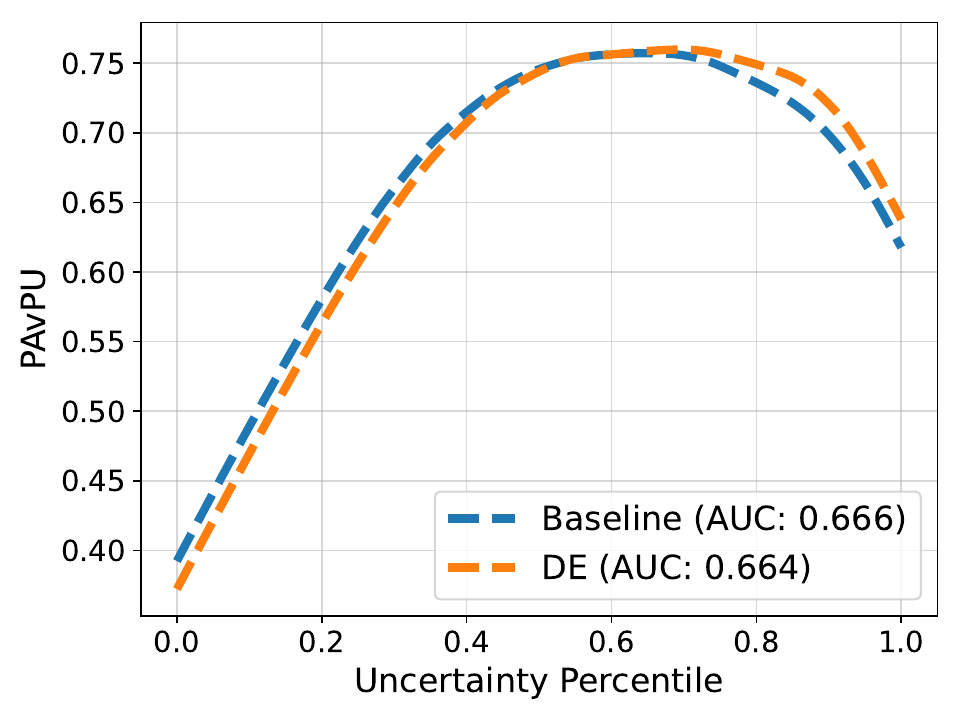}
        \caption{Depth: PAvPU}
        \label{fig: subfig6}
    \end{subfigure}
    \caption{Out-of-Domain (OOD) uncertainty quality evaluation between the baseline SegDepthFormer and a SegDepthFormer DE with 10 members on the Rain$_3$ Cityscapes validation dataset \cite{hu2019depth}. Rain$_3$ uses [0.03, 0.015, 0.002] for attenuation coefficients $\alpha$ and $\beta$ and the raindrop radius $a$. We compare the three uncertainty metrics $p(\text{accurate}|\text{certain})$, $p(\text{uncertain}|\text{inaccurate})$, and PAvPU for different uncertainty thresholds. Additionally, we report the area under curve (AUC).}
    \label{fig:OOD}
\end{figure}

\textbf{Uncertainty Quality.} To account for the impact of the uncertainty threshold on uncertainty quality metrics, as discussed in Section \ref{sec: UT Impact}, we assess the OOD uncertainty quality across the entire percentile spectrum and utilize the area under the curve (AUC) as a comprehensive evaluation metric. Figure \ref{fig:OOD} provides a detailed comparison between the SegDepthFormer baseline and the DE on the Rain$_3$ Cityscapes validation dataset. The DE demonstrates consistent superiority over the baseline in $p(\text{accurate}|\text{certain})$ (AUC: 0.931 vs. 0.949 for segmentation and 0.847 vs. 0.861 for depth) and $p(\text{uncertain}|\text{inaccurate})$ (AUC: 0.759 vs. 0.794 for segmentation and 0.717 vs. 0.726 for depth). For PAvPU, the DE performs comparably to the baseline (AUC: 0.597 vs. 0.602 for segmentation and 0.666 vs. 0.664 for depth), aligning with the findings in Table \ref{table:uq-rain}. 

\section{Conclusion}
By comparing uncertainty quantification methods in joint semantic segmentation and monocular depth estimation, we find Deep Ensembles offer the best performance and uncertainty quality, albeit at higher computational cost. Deep Sub-Ensembles provide an efficient alternative with minimal trade-offs in predictive performance and uncertainty quality. Additionally, we reveal that multi-task learning can enhance the uncertainty quality of semantic segmentation compared to solving both tasks separately. Furthermore, we show that while the choice of the uncertainty threshold significantly impacts metrics, its influence remains independent of the underlying model or approach and the median uncertainty of an image proves to be a suitable default threshold with high values for $p(\text{accurate}|\text{certain})$ and $p(\text{uncertain}|\text{inaccurate})$. Lastly, we find that Deep Ensembles exhibit robustness in out-of-domain scenarios, offering superior predictive performance and uncertainty quality.

\begin{acknowledgement}
The authors acknowledge support by the state of Baden-Württemberg through bwHPC.\\
This work is supported by the Helmholtz Association Initiative and Networking Fund on the HAICORE@KIT partition.\\
\end{acknowledgement}

\bibliographystyle{ieee_fullname}
\bibliography{bibliography}

\begin{thebibliography}{10}\itemsep=-1pt

\bibitem{bruggemann2020automated}
David Bruggemann, Menelaos Kanakis, Stamatios Georgoulis, and Luc Van~Gool.
\newblock Automated search for resource-efficient branched multi-task networks.
\newblock {\em arXiv preprint arXiv:2008.10292}, 2020.

\bibitem{bruggemann2021exploring}
David Br{\"u}ggemann, Menelaos Kanakis, Anton Obukhov, Stamatios Georgoulis, and Luc Van~Gool.
\newblock Exploring relational context for multi-task dense prediction.
\newblock In {\em Proceedings of the IEEE/CVF international conference on computer vision}, pages 15869--15878, 2021.

\bibitem{cordts2016CityscapesDataset}
Marius Cordts, Mohamed Omran, Sebastian Ramos, Timo Rehfeld, Markus Enzweiler, Rodrigo Benenson, Uwe Franke, Stefan Roth, and Bernt Schiele.
\newblock The cityscapes dataset for semantic urban scene understanding.
\newblock In {\em Proceedings of the IEEE Conference on Computer Vision and Pattern Recognition (CVPR)}, 2016.

\bibitem{deng2009ImageNetLargescale}
Jia Deng, Wei Dong, Richard Socher, Li-Jia Li, {Kai Li}, and {Li Fei-Fei}.
\newblock {{ImageNet}}: {{A}} large-scale hierarchical image database.
\newblock In {\em 2009 {{IEEE Conference}} on {{Computer Vision}} and {{Pattern Recognition}}}, pages 248--255. {IEEE}, 2009.

\bibitem{dong2022towards}
Xingshuai Dong, Matthew~A Garratt, Sreenatha~G Anavatti, and Hussein~A Abbass.
\newblock Towards real-time monocular depth estimation for robotics: A survey.
\newblock {\em IEEE Transactions on Intelligent Transportation Systems}, 23(10):16940--16961, 2022.

\bibitem{fort2020DeepEnsembles}
Stanislav Fort, Huiyi Hu, and Balaji Lakshminarayanan.
\newblock Deep {{Ensembles}}: {{A Loss Landscape Perspective}}.
\newblock {\em arXiv:1912.02757}, 2020.

\bibitem{gal2016uncertainty}
Yarin Gal.
\newblock Uncertainty in deep learning.
\newblock {\em Ph.D. thesis, University of Cambridge}, 2016.

\bibitem{gal2016DropoutBayesian}
Yarin Gal and Zoubin Ghahramani.
\newblock Dropout as a bayesian approximation: Representing model uncertainty in deep learning.
\newblock In {\em Proceedings of The 33rd International Conference on Machine Learning}, volume~48 of {\em Proceedings of Machine Learning Research}, pages 1050--1059. PMLR, 2016.

\bibitem{gal2017deep}
Yarin Gal, Riashat Islam, and Zoubin Ghahramani.
\newblock Deep bayesian active learning with image data.
\newblock In {\em International conference on machine learning}, pages 1183--1192. PMLR, 2017.

\bibitem{gao2022ci}
Tianxiao Gao, Wu Wei, Zhongbin Cai, Zhun Fan, Sheng~Quan Xie, Xinmei Wang, and Qiuda Yu.
\newblock Ci-net: A joint depth estimation and semantic segmentation network using contextual information.
\newblock {\em Applied Intelligence}, 52(15):18167--18186, 2022.

\bibitem{gawlikowski2022SurveyUncertainty}
Jakob Gawlikowski, Cedrique Rovile~Njieutcheu Tassi, Mohsin Ali, Jongseok Lee, Matthias Humt, Jianxiang Feng, Anna Kruspe, Rudolph Triebel, Peter Jung, Ribana Roscher, Muhammad Shahzad, Wen Yang, Richard Bamler, and Xiao~Xiang Zhu.
\newblock A {{Survey}} of {{Uncertainty}} in {{Deep Neural Networks}}.
\newblock {\em arXiv:2107.03342}, 2022.

\bibitem{guo2017CalibrationModerna}
Chuan Guo, Geoff Pleiss, Yu Sun, and Kilian~Q. Weinberger.
\newblock On calibration of modern neural networks.
\newblock In {\em Proceedings of the 34th International Conference on Machine Learning}, pages 1321--1330. PMLR, 2017.

\bibitem{gustafsson2020evaluating}
Fredrik~K Gustafsson, Martin Danelljan, and Thomas~B Schon.
\newblock Evaluating scalable bayesian deep learning methods for robust computer vision.
\newblock In {\em Proceedings of the IEEE/CVF conference on computer vision and pattern recognition workshops}, pages 318--319, 2020.

\bibitem{he2021sosd}
Lei He, Jiwen Lu, Guanghui Wang, Shiyu Song, and Jie Zhou.
\newblock Sosd-net: Joint semantic object segmentation and depth estimation from monocular images.
\newblock {\em Neurocomputing}, 440:251--263, 2021.

\bibitem{hu2019depth}
Xiaowei Hu, Chi-Wing Fu, Lei Zhu, and Pheng-Ann Heng.
\newblock Depth-attentional features for single-image rain removal.
\newblock In {\em Proceedings of the IEEE/CVF Conference on computer vision and pattern recognition}, pages 8022--8031, 2019.

\bibitem{ji2023semantic}
Naihua Ji, Huiqian Dong, Fanyun Meng, and Liping Pang.
\newblock Semantic segmentation and depth estimation based on residual attention mechanism.
\newblock {\em Sensors}, 23(17):7466, 2023.

\bibitem{jiao2018look}
Jianbo Jiao, Ying Cao, Yibing Song, and Rynson Lau.
\newblock Look deeper into depth: Monocular depth estimation with semantic booster and attention-driven loss.
\newblock In {\em Proceedings of the European conference on computer vision (ECCV)}, pages 53--69, 2018.

\bibitem{kendall2017CVUncertainties}
Alex Kendall and Yarin Gal.
\newblock What uncertainties do we need in bayesian deep learning for computer vision?
\newblock In {\em Proceedings of the 31st International Conference on Neural Information Processing Systems}, page 5580–5590, 2017.

\bibitem{kendall2018multi}
Alex Kendall, Yarin Gal, and Roberto Cipolla.
\newblock Multi-task learning using uncertainty to weigh losses for scene geometry and semantics.
\newblock In {\em Proceedings of the IEEE conference on computer vision and pattern recognition}, pages 7482--7491, 2018.

\bibitem{lakshminarayanan2017SimpleScalable}
Balaji Lakshminarayanan, Alexander Pritzel, and Charles Blundell.
\newblock Simple and scalable predictive uncertainty estimation using deep ensembles.
\newblock In {\em Advances in Neural Information Processing Systems}, volume~30. {Curran Associates, Inc.}, 2017.

\bibitem{landgraf2024evaluation}
Steven Landgraf, Markus Hilleman, Theodor Kapler, and Markus Ulrich.
\newblock Evaluation of multi-task uncertainties in joint semantic segmentation and monocular depth estimation.
\newblock In {\em Forum Bildverarbeitung 2024}, page 147. KIT Scientific Publishing, 2024.

\bibitem{landgraf2023segmentation}
S Landgraf, M Hillemann, M Aberle, V Jung, and M Ulrich.
\newblock Segmentation of industrial burner flames: A comparative study from traditional image processing to machine and deep learning.
\newblock {\em ISPRS Annals of Photogrammetry, Remote Sensing \& Spatial Information Sciences}, 10, 2023.

\bibitem{landgraf2024efficient}
Steven Landgraf, Markus Hillemann, Theodor Kapler, and Markus Ulrich.
\newblock Efficient multi-task uncertainties for joint semantic segmentation and monocular depth estimation.
\newblock In {\em DAGM German Conference on Pattern Recognition (GCPR)}. Springer, 2024.

\bibitem{landgraf2024uce}
S. Landgraf, M. Hillemann, K. Wursthorn, and M. Ulrich.
\newblock Uncertainty-aware cross-entropy for semantic segmentation.
\newblock {\em ISPRS Annals of the Photogrammetry, Remote Sensing and Spatial Information Sciences}, X-2, 2024.

\bibitem{landgraf2023dudes}
Steven Landgraf, Kira Wursthorn, Markus Hillemann, and Markus Ulrich.
\newblock Dudes: Deep uncertainty distillation using ensembles for semantic segmentation.
\newblock {\em PFG--Journal of Photogrammetry, Remote Sensing and Geoinformation Science}, 92(2):101--114, 2024.

\bibitem{lee2018TrainingConfidencecalibrated}
Kimin Lee, Honglak Lee, Kibok Lee, and Jinwoo Shin.
\newblock Training {{Confidence-calibrated Classifiers}} for {{Detecting Out-of-Distribution Samples}}.
\newblock {\em arXiv:1711.09325}, 2018.

\bibitem{lin2019depth}
Xiao Lin, Dalila S{\'a}nchez-Escobedo, Josep~R Casas, and Montse Pard{\`a}s.
\newblock Depth estimation and semantic segmentation from a single rgb image using a hybrid convolutional neural network.
\newblock {\em Sensors}, 19(8):1795, 2019.

\bibitem{liu2018collaborative}
Jing Liu, Yuhang Wang, Yong Li, Jun Fu, Jiangyun Li, and Hanqing Lu.
\newblock Collaborative deconvolutional neural networks for joint depth estimation and semantic segmentation.
\newblock {\em IEEE transactions on neural networks and learning systems}, 29(11):5655--5666, 2018.

\bibitem{liu2019end}
Shikun Liu, Edward Johns, and Andrew~J Davison.
\newblock End-to-end multi-task learning with attention.
\newblock In {\em Proceedings of the IEEE/CVF conference on computer vision and pattern recognition}, pages 1871--1880, 2019.

\bibitem{loquercio2020general}
Antonio Loquercio, Mattia Segu, and Davide Scaramuzza.
\newblock A general framework for uncertainty estimation in deep learning.
\newblock {\em IEEE Robotics and Automation Letters}, 5(2):3153--3160, 2020.

\bibitem{loshchilov2017decoupled}
Ilya Loshchilov and Frank Hutter.
\newblock Decoupled weight decay regularization.
\newblock {\em arXiv preprint arXiv:1711.05101}, 2017.

\bibitem{mackay1992PracticalBayesian}
David J.~C. MacKay.
\newblock A {{Practical Bayesian Framework}} for {{Backpropagation Networks}}.
\newblock {\em Neural Computation}, 4(3):448--472, 1992.

\bibitem{mcallister2017ConcreteProblems}
Rowan McAllister, Yarin Gal, Alex Kendall, Mark {van der Wilk}, Amar Shah, Roberto Cipolla, and Adrian Weller.
\newblock Concrete {{Problems}} for {{Autonomous Vehicle Safety}}: {{Advantages}} of {{Bayesian Deep Learning}}.
\newblock In {\em Proceedings of the {{Twenty-Sixth International Joint Conference}} on {{Artificial Intelligence}}}, pages 4745--4753, 2017.

\bibitem{minaee2020ImageSegmentation}
Shervin Minaee, Yuri Boykov, Fatih Porikli, Antonio Plaza, Nasser Kehtarnavaz, and Demetri Terzopoulos.
\newblock Image segmentation using deep learning: A survey.
\newblock {\em IEEE Transactions on Pattern Analysis and Machine Intelligence}, 44(7):3523--3542, 2022.

\bibitem{mousavian2016joint}
Arsalan Mousavian, Hamed Pirsiavash, and Jana Ko{\v{s}}eck{\'a}.
\newblock Joint semantic segmentation and depth estimation with deep convolutional networks.
\newblock In {\em 2016 Fourth International Conference on 3D Vision (3DV)}, pages 611--619. IEEE, 2016.

\bibitem{mukhoti2018evaluating}
Jishnu Mukhoti and Yarin Gal.
\newblock Evaluating bayesian deep learning methods for semantic segmentation.
\newblock {\em arXiv preprint arXiv:1811.12709}, 2018.

\bibitem{naeini2015obtaining}
Mahdi~Pakdaman Naeini, Gregory Cooper, and Milos Hauskrecht.
\newblock Obtaining well calibrated probabilities using bayesian binning.
\newblock In {\em Proceedings of the AAAI conference on artificial intelligence}, volume~29, 2015.

\bibitem{nekrasov2019real}
Vladimir Nekrasov, Thanuja Dharmasiri, Andrew Spek, Tom Drummond, Chunhua Shen, and Ian Reid.
\newblock Real-time joint semantic segmentation and depth estimation using asymmetric annotations.
\newblock In {\em 2019 International Conference on Robotics and Automation (ICRA)}, pages 7101--7107. IEEE, 2019.

\bibitem{nix1994estimating}
David~A Nix and Andreas~S Weigend.
\newblock Estimating the mean and variance of the target probability distribution.
\newblock In {\em Proceedings of 1994 ieee international conference on neural networks (ICNN'94)}, volume~1, pages 55--60. IEEE, 1994.

\bibitem{sakaridis2018semantic}
Christos Sakaridis, Dengxin Dai, and Luc Van~Gool.
\newblock Semantic foggy scene understanding with synthetic data.
\newblock {\em International Journal of Computer Vision}, 126:973--992, 2018.

\bibitem{silberman2012indoor}
Nathan Silberman, Derek Hoiem, Pushmeet Kohli, and Rob Fergus.
\newblock Indoor segmentation and support inference from rgbd images.
\newblock In {\em Computer Vision--ECCV 2012: 12th European Conference on Computer Vision, 2012, Proceedings, Part V 12}, pages 746--760. Springer, 2012.

\bibitem{steger2018MachineVision}
Carsten Steger, Markus Ulrich, and Christian Wiedemann.
\newblock {\em Machine Vision Algorithms and Applications}.
\newblock John Wiley \& Sons, 2018.

\bibitem{valdenegro2023sub}
Matias Valdenegro-Toro.
\newblock Sub-ensembles for fast uncertainty estimation in neural networks.
\newblock In {\em Proceedings of the IEEE/CVF International Conference on Computer Vision}, pages 4119--4127, 2023.

\bibitem{wang2015towards}
Peng Wang, Xiaohui Shen, Zhe Lin, Scott Cohen, Brian Price, and Alan~L Yuille.
\newblock Towards unified depth and semantic prediction from a single image.
\newblock In {\em Proceedings of the IEEE conference on computer vision and pattern recognition}, pages 2800--2809, 2015.

\bibitem{wolf2024decoupling}
Dominik~Werner Wolf, Prasannavenkatesh Balaji, Alexander Braun, and Markus Ulrich.
\newblock Decoupling of neural network calibration measures.
\newblock In {\em DAGM German Conference on Pattern Recognition}. Springer, 2024.

\bibitem{wursthorn2024uq}
K. Wursthorn, M. Hillemann, and M. Ulrich.
\newblock Uncertainty quantification with deep ensembles for 6d object pose estimation.
\newblock {\em ISPRS Annals of the Photogrammetry, Remote Sensing and Spatial Information Sciences}, X-2, 2024.

\bibitem{xie2021segformer}
Enze Xie, Wenhai Wang, Zhiding Yu, Anima Anandkumar, Jose~M Alvarez, and Ping Luo.
\newblock Segformer: Simple and efficient design for semantic segmentation with transformers.
\newblock {\em Advances in Neural Information Processing Systems}, 34:12077--12090, 2021.

\bibitem{xu2018pad}
Dan Xu, Wanli Ouyang, Xiaogang Wang, and Nicu Sebe.
\newblock Pad-net: Multi-tasks guided prediction-and-distillation network for simultaneous depth estimation and scene parsing.
\newblock In {\em Proceedings of the IEEE Conference on Computer Vision and Pattern Recognition}, pages 675--684, 2018.

\bibitem{xu2022mtformer}
Xiaogang Xu, Hengshuang Zhao, Vibhav Vineet, Ser-Nam Lim, and Antonio Torralba.
\newblock Mtformer: Multi-task learning via transformer and cross-task reasoning.
\newblock In {\em European Conference on Computer Vision}, pages 304--321. Springer, 2022.

\end{thebibliography}

\end{document}